\documentclass{article}
\usepackage{spconf,amsmath,graphicx}
\usepackage{bm}
\usepackage{color}
\usepackage{url}
\usepackage{here}
\usepackage{mathtools}
\usepackage{multirow}
\usepackage{amsmath,amssymb}

\DeclarePairedDelimiter\floor{\lfloor}{\rfloor}

\title{Minimum Latency Training Strategies\\for Streaming Sequence-to-Sequence ASR}
\name{Hirofumi Inaguma$^{12}$$^\dagger$\thanks{$^\dagger$Work performed during an internship at Microsoft.}, Yashesh Gaur$^2$, Liang Lu$^2$, Jinyu Li$^2$, Yifan Gong$^2$}
\address{$^1$Kyoto University, Kyoto, Japan, $^2$Microsoft Speech and Language Group, Redmond, WA, USA}
\begin{document}
\ninept
\maketitle
\begin{abstract}
Recently, a few novel streaming attention-based sequence-to-sequence (S2S) models have been proposed to perform online speech recognition with linear-time decoding complexity.
However, in these models, the decisions to generate tokens are delayed compared to the actual acoustic boundaries since their unidirectional encoders lack future information.
This leads to an inevitable latency during inference.
To alleviate this issue and reduce latency, we propose several strategies during training by leveraging external hard alignments extracted from the hybrid model.
We investigate to utilize the alignments in both the encoder and the decoder.
On the encoder side, (1) multi-task learning and (2) pre-training with the framewise classification task are studied.
On the decoder side, we (3) remove inappropriate alignment paths beyond an acceptable latency during the alignment marginalization, and (4) directly minimize the differentiable expected latency loss.
% marginalization of all possible alignments
Experiments on the Cortana voice search task demonstrate that our proposed methods can significantly reduce the latency, and even improve the recognition accuracy in certain cases on the decoder side.
% with 3.4k hours training data
We also present some analysis to understand the behaviors of streaming S2S models.
% streaming attention-based S2S
\end{abstract}
%
%%%%%%%%%%%%%%%%%%%%%%%%%%%%%%%%%%%%%%%%%%%%%%%%%%
% Introduction
%%%%%%%%%%%%%%%%%%%%%%%%%%%%%%%%%%%%%%%%%%%%%%%%%%
\section{Introduction}\label{sec:intro}
% end-to-end ASR
End-to-end automatic speech recognition (ASR) systems have been successfully developed and nowadays achieve the competitive performance of conventional hybrid systems \cite{google_sota_asr}.
Among various end-to-end ASR models, attention-based sequence-to-sequence (S2S) models \cite{las,chorowski2015attention}, have been shown to perform superior \cite{s2s_comparison_google,s2s_comparison_baidu} to other models like connectionist temporal classification (CTC) \cite{ctc_graves,graves2014towards} and recurrent neural transducer (RNN-T) \cite{rnnt}.
However, unlike frame-synchronous models such as CTC and RNN-T, it is difficult to directly apply S2S models to the streaming scenario because of global attention score normalization over all the encoder memories.
% memory chunks

\begin{figure}[t]
    \centering
    \vspace{-2mm}
    \includegraphics[width=0.9\hsize]{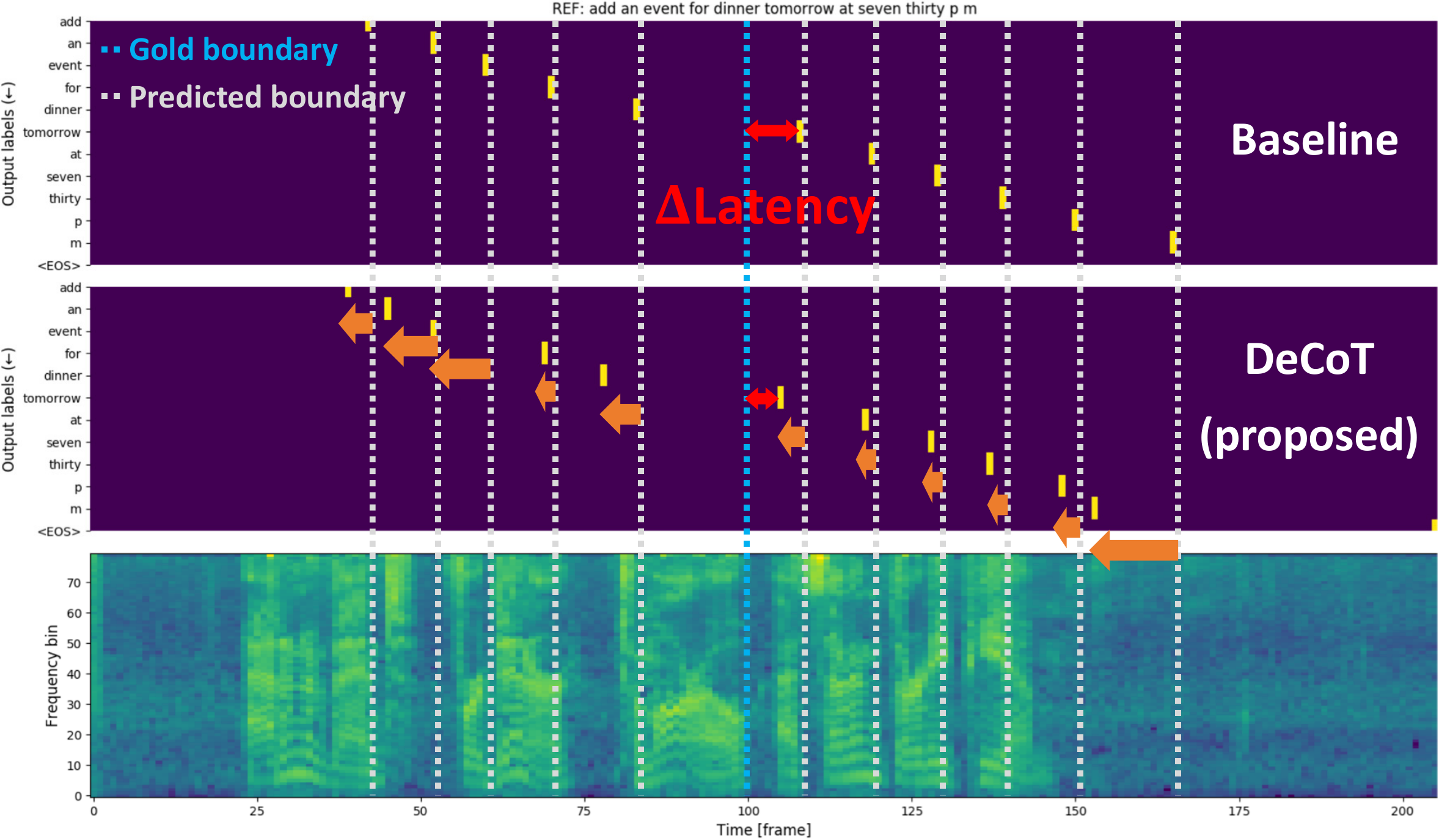}
    \vspace{-3mm}
    \caption{Monotonic attention weights $\{\alpha_{i,j}\}$ \textit{w.r.t.} MoChA during the inference. Yellow dots represent predicted decision boundaries for the token generation.}
    \label{fig:delayed_prediction}
    \vspace{-5mm}
\end{figure}

% streaming ASR, delayed prediction problem
To address this, several novel streaming attention-based S2S models have been proposed including local windowing methods \cite{luong2015effective,chan2016online,tjandra2017local,hou2017gaussian,attention_variants}, neural transducer \cite{neural_transducer}, hard monotonic attention \cite{monotonic_attention,mocha}, adaptive computation steps \cite{adaptive_computation_steps}, triggered attention \cite{triggered_attention}, and continuous integrate-and-fire \cite{cif}.
% add more references?
% These frameworks are well-designed and show promising results.
Among these well-designed frameworks, monotonic chunkwise attention (MoChA) \cite{mocha} can be optimized efficiently in parallel and shows promising results on the large-scale ASR task \cite{online_hybrid_ctc_attention,adaptive_mocha,kim2020attention,garg2019improved}.
However, these `streaming' models are still unusable in many online tasks as their latency regarding the token generation is not small enough.
This is an important issue that current research has not addressed well.  
We have observed that the decision boundary on the token generation for these models is delayed from the actual acoustic boundary (see Figure \ref{fig:delayed_prediction}).
This is because the unidirectional encoder lacks future information, and the model is optimized to utilize as many future frames as possible to maximize the log probabilities over the transcription \cite{sak2015learning_icassp}.
This leads to inevitable latency and hurts user experiences.
% This problem has been tackled in the CTC framework \cite{sak2015acoustic_asru,dfsmn_ctc} by leveraging the hard alignments as supervision.
% However, since streaming attention-based S2S models like MoChA are not frame-synchronous, it is not easy to apply framewise supervision to them.

% proposed methods
In this paper, we propose minimum latency training strategies to reduce the latency of the streaming attention-based S2S models.
To the best of our knowledge, this has not been investigated for such ASR models so far.
In this work, we focus on the algorithmic delay regarding the token generation and refer it to \textit{latency}.
Inspired by \cite{sak2015acoustic_asru,dfsmn_ctc}, we leverage external hard alignments extracted from a hybrid model as supervision.
The goal is to force the model to learn accurate alignments, which would reduce the latency, while keeping the recognition accuracy.
We adopt MoChA as a streaming S2S model and explore to utilize the alignments both on the encoder and decoder sides.
% a streaming attention-based S2S model
On the encoder side, we perform (1) multi-task learning and (2) pre-training with the framewise cross-entropy objective, which have shown to be effective to stabilize CTC training and further improve the ASR accuracy \cite{sak2015learning_icassp,dfsmn_ctc,multistage_a2w}.
On the decoder side, we propose two novel methods: (3) delay constrained training (DeCoT) and (4) minimum latency training (MinLT).
DeCoT is conducted by removing inappropriate alignments during the marginalization process of all possible alignments \cite{sak2015acoustic_asru}.
Moreover, a regularization term is introduced to avoid the exponential decay of attention weights because of the sequential dependency in the decoder.
MinLT is performed by directly minimizing a differentiable expected latency estimated from the expected boundary locations.

% results
Experimental evaluations on Cortana 3.4k hours dataset demonstrate that our proposed methods significantly reduce the latency in both cases, and DeCoT and MinLT are more effective.
Surprisingly, we confirm significant improvements of the ASR accuracy as well when leveraging the alignment information on the decoder side.
Furthermore, the ablation study is conducted to understand the behaviors of streaming S2S models.

% This paper is organized as follows: In Section 2, we describe our streaming S2S model. 
% In Section 3, we define the evaluation metric for latency. 
% In Section 4, we propose training strategies to reduce the latency by using the hard alignments.
% We conduct experiments in Section 5 and conclude this paper in Section 6.

% The latency is defined as the difference between the predicted boundary (yellow dots) and the corresponding reference boundary (red dot line).
% the predicted boundary (yellow dot) in the top row is delayed from the corresponding acoustic boundary (blue line).

%%%%%%%%%%%%%%%%%%%%%%%%%%%%%%%%%%%%%%%%%%%%%%%%%%
% Baseline method
%%%%%%%%%%%%%%%%%%%%%%%%%%%%%%%%%%%%%%%%%%%%%%%%%%
\section{Streaming sequence-to-sequence ASR}\label{sec:streaming}
\vspace{-2mm}
\subsection{Monotonic chunkwise attention (MoChA)}\label{ssec:mocha}
We build streaming attention-based sequence-to-sequence (S2S) models based on the monotonic chunkwise attention (MoChA) model \cite{mocha}.
MoChA is an extension of the hard monotonic attention model \cite{monotonic_attention} and introduces additional soft chunkwise attention mechanism on top of it.
During training, monotonic alignments are learnt by marginalizing over all possible alignments $\{\alpha_{i,j}\}$ represented by selection probabilities $\{p_{i,j}\}$ as follows:
% monotonic attention
\vspace{-1mm}
\begin{eqnarray}
e^{\rm mono}_{i,j} &=& g\frac{v^{\rm T}}{||v||}{\rm ReLU}({\mathbf W}_{\rm h}h_{j} + {\mathbf W}_{\rm s}s_{i} + b) + r \\ \label{eq:mocha_energy}
% e^{\rm mono}_{i,j} &=& {\rm MonotonicAttention}(s_{i}, h_{j}) \\ \label{eq:mocha_energy}
p_{i,j} &=& \sigma(e^{\rm mono}_{i,j}) \nonumber \\
\alpha_{i,j} &=& p_{i,j}\sum_{k=1}^{j}\bigg(\alpha_{i-1,k}\prod_{l=k}^{j-1}(1-p_{i,l})\bigg) \nonumber \\
&=& p_{i,j}\bigg((1-p_{i,j-1})\frac{\alpha_{i,j-1}}{p_{i,j-1}}+\alpha_{i-1,j}\bigg) \label{eq:mocha_alpha}
\end{eqnarray}
where $g$, $v$, ${\mathbf W}_{\rm h}$, ${\mathbf W}_{\rm s}$, $b$, and $r$ are learnable parameters, $s_{i}$ is the $i$-th decoder state, $h_{j}$ is the $j$-th encoder state, $\sigma$ is a logistic sigmoid function, and $e_{i,j}^{\rm mono}$ is the monotonic energy activation.
Additional soft chunkwise attention is performed over small chunks ($w$ frames) from each boundary by normalizing the chunk energy activation $e_{i,j}^{\rm chunk}$ (implemented with different parameters as in Eq. \eqref{eq:mocha_energy}):
% chunkwise attention
\vspace{-1mm}
\begin{equation}\label{eq:mocha_beta}
\beta_{i,j} = \sum_{k=j}^{j+w-1}\bigg(\alpha_{i,k}{\rm exp}(e_{i,j}^{\rm chunk})/\sum_{l=k-w+1}^{k}{{\rm exp}(e_{i,l}^{\rm chunk})}\bigg)
\end{equation}
The expected context vector is calculated as a weighted encoder memories by $c_{i} = \sum_{j=1}^{T}\beta_{i,j}h_{j}$ and the following token generation processes are the same as the global attention.
% mechanism
Fortunately, Eq. \eqref{eq:mocha_alpha} and \eqref{eq:mocha_beta} can be calculated efficiently in parallel with the cumulative sum and product, and moving sum operations during training.
At test time, each token is generated once $p_{i,j}$ surpasses a threshold 0.5 and then $\alpha_{i,j}$ is set to 1.0.
See \cite{monotonic_attention,mocha} for more details.
% \begin{eqnarray*}\label{eq:mocha_parallel}
% {\bm \alpha}_{i}={\bm p}_{i}\cdot{\rm cumprod}(1-{\bm p}_{i})\cdot{\rm cumsum}\bigg(\frac{{\bm \alpha}_{i-1}}{{\rm cumprod}(1-{\bm p}_{i})}\bigg)
% \end{eqnarray*}
% Note that we do not utilize the auxiliary CTC objective \cite{triggered_attention,online_hybrid_ctc_attention} because we are interested in the modeling capability of the streaming S2S model itself.

\vspace{-1mm}
\subsection{Enhanced monotonic attention with 1D convolution}
Although monotonic alignments are efficiently formulated in Eq. \eqref{eq:mocha_alpha}, its binary decision on whether to generate the next token is parameterized by $p_{i,j}$ and it depends on the corresponding encoder output $h_{j}$.
In theory, RNN encoder should capture past information over several timesteps.
However, we found that using surrounding encoder outputs as additional key features in Eq. \eqref{eq:mocha_energy} is effective for the robust binary decision.
Specifically, we introduce 1-dimensional convolutional layer ${\mathbf W}_{\rm c} \in \mathbb{R}^{k \times k \times d}$ before transforming $h_{j}$ into the attention space by $h'_{i,j}={\mathbf W}_{\rm h}({\mathbf W}_{\rm c}*h_{j})$.
We set the kernel size $k$ to 5 and channel size $d$ to the dimension of the encoder output.
Note that this enhanced MoChA decoder looks at $\floor*{k/2-1}$ ($=2$) future frames at every input timestep $j$ for the boundary prediction.

%%%%%%%%%%%%%%%%%%%%%%%%%%%%%%%%%%%%%%%%%%%%%%%%%%
% Problem specification
%%%%%%%%%%%%%%%%%%%%%%%%%%%%%%%%%%%%%%%%%%%%%%%%%%
\section{Problem specification}\label{ssec:problem}
\vspace{-2mm}
\subsection{Definition of latency}
S2S models do not guarantee that token boundaries are aligned to the corresponding acoustic boundaries accurately \cite{sak2015learning_icassp}.
When the unidirectional encoder is used, monotonic attention weights $\{\alpha_{i,j}\}$ are generally distributed several future frames ahead to maximize the log probabilities of the target sequence as much as possible, which causes the inevitable latency \cite{kim2020attention} (see Figure \ref{fig:delayed_prediction}).

In this work, we focus on this issue and define this delay as \textit{latency}.
The goal of this work is to reduce the latency as much as possible while maintaining the recognition accuracy.
The most relevant works are \cite{sak2015acoustic_asru,dfsmn_ctc}, where the authors also tackle the same issue in the CTC acoustic model.
However, to the best of our knowledge, this problem has not been investigated so far in the streaming attention-based S2S models, which are categorized as label-synchronous models and behave very differently from frame-synchronous models.

\vspace{-2mm}
\subsection{Evaluation metric}
In this work, we adopt \textit{corpus-level} latency $\Delta_{\rm corpus}$ and \textit{utterance-level} latency $\Delta_{\rm utterance}$ as evaluation metrics for latency.
We regard input timesteps where monotonic attention weights are activated ($\{\alpha_{i,j}|\alpha_{i,j}=1\}_{i=1,\ldots,L}$) as the predicted boundaries.
Latency of each token in the $k$-th utterance is calculated as the difference between the predicted boundary $\hat{\rm{b}}_{i}^{k}$ and the corresponding gold boundary ${\rm b}_{i}^{k}$.
For the corpus-level latency, we take an average of all tokens in the evaluation set as follows:
% corpus-level latency
\vspace{-2mm}
\[
\Delta_{\rm corpus} = \frac{1}{\sum_{k=1}^{N}{|\bm{y}^{k}|}}\sum_{k=1}^{N}\sum_{i=1}^{|\bm{y}^{k}|}(\hat{\rm{b}}_{i}^{k} - {\rm b}_{i}^{k})
\]
where $N$ is the number of utterances in the evaluation set and $\bm{y}^{k}$ is the $k$-th reference.
For the utterance-level latency, we take an average of the mean latency in each utterance as follows:
% utterance-level latency
\vspace{-2mm}
\[
\Delta_{\rm utterance} = \frac{1}{N}\sum_{k=1}^{N}\frac{1}{|\bm{y}^{k}|}\sum_{i=1}^{|\bm{y}^{k}|}(\hat{\rm{b}}_{i}^{k} - {\rm b}_{i}^{k})
\]
We mainly report average, median, 90th and 99th percentile of the corpus-level latency distributions.
Since the length of hypothesis must be equal to that of the corresponding reference for these metrics, we conduct teacher-forcing when calculating the latency.
% we conduct decoding in the teacher-forcing mode (as in the training stage) when calculating the latency.

\vspace{-1mm}
%%%%%%%%%%%%%%%%%%%%%%%%%%%%%%%%%%%%%%%%%%%%%%%%%%
% Proposed method
%%%%%%%%%%%%%%%%%%%%%%%%%%%%%%%%%%%%%%%%%%%%%%%%%%
\section{Strategies for minimum latency training}\label{sec:strategies}
\vspace{-1mm}
In this section, we propose minimum latency training strategies applicable in the encoder and decoder to reduce the latency of streaming S2S models.
We leverage external hard alignments obtained from the acoustic model in the hybrid system.
%%%%%%%%%%%%%%%%%%%%%%%%%%%%%%%%%%%%%%%%%%%%%%%%
% describe the acoustic model in detail here
%%%%%%%%%%%%%%%%%%%%%%%%%%%%%%%%%%%%%%%%%%%%%%%%
The acoustic model is trained to minimize senone-level framewise cross-entropy (CE) loss.
Let ${\bm A}=(a_{1}, \cdots, a_{T})$ ($a_{j}$ is a $K$-dimensional one-hot vector, $K$: vocabulary size) be a hard alignment corresponding to the input sequence ${\bm x}=(x_{1}, \cdots, x_{T})$, and ${\bm b}=(b_{1}, \cdots, b_{L})$ be a sequence of token boundaries (end points) for each reference ${\bm y}=(y_{1}, \cdots, y_{L})$.
${\bm b}$ can be obtained from ${\bm A}$.

\vspace{-2mm}
\subsection{Leveraging hard alignments in the encoder}
\vspace{-1mm}
%%% MTL with framewise CE
\subsubsection{Multi-task learning with framewise CE objective (MTL-CE)}\label{ssec:mtl_frame_xe}
\begin{figure}[!t]
    \centering
    \includegraphics[width=0.95\hsize]{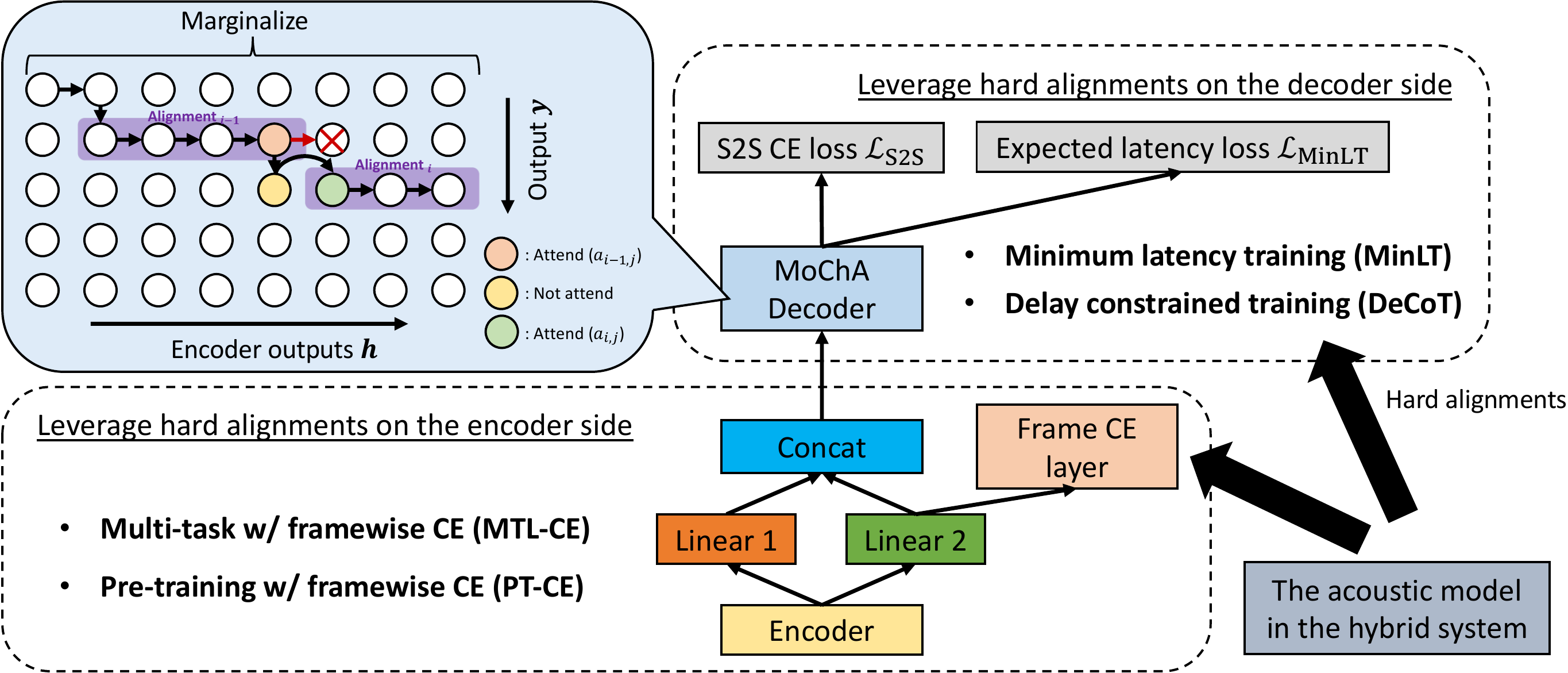}
    \vspace{-3mm}
    \caption{System overview of leveraging external hard alignments}
    \label{fig:system_overview}
    \vspace{-4mm}
\end{figure}

We first propose the multi-task learning with framewise cross-entropy (CE) objective by using the hard alignments.
We hypothesize that framewise supervision regularize encoder representations so that each encoder output $h_{j}$ is aligned to the true acoustic location, which would be helpful for calculating accurate boundaries in Eq. \eqref{eq:mocha_energy}.
We attach another softmax layer for framewise CE objective on top of the encoder and jointly optimize CE objective $\mathcal{L}_{\rm S2S}$ (\textit{S2S branch}) and framewise CE objective $\mathcal{L}_{\rm CE}$ (\textit{CE branch}) by linearly interpolating with a tunable hyperparameter $\lambda_{\rm CE}$ ($0 \le \lambda_{\rm CE} \le 1$):
\begin{equation}\label{eq:mtl}
\mathcal{L}_{\rm total}=(1 - \lambda_{\rm CE}) \mathcal{L}_{\rm S2S}({\bm y}|{\bm x}) + \lambda_{\rm CE} \mathcal{L}_{\rm CE}({\bm A}|{\bm x})
\end{equation}
% $\mathcal{L}_{\rm S2S}({\bm A}|{\bm x})=-\sum_{i=1}^{L}y_{i}\log{q_{j}^{\rm S2S}}$, 
where $\mathcal{L}_{\rm CE}({\bm A}|{\bm x})=-\sum_{j=1}^{T}a_{j}\log{q_{j}^{\rm CE}}$, and $q_{j}^{\rm CE}$ are posterior distributions from the CE branch.
Following \cite{multistage_a2w}, we insert two linear projection layers after the top encoder layer for each branch as the bottleneck layers (see Figure \ref{fig:system_overview}).
Both outputs from two projection layers are concatenated and fed into the S2S branch.
The softmax layer in the framewise CE branch is discarded during inference.

\vspace{-1mm}
%%% CE initialization
\subsubsection{Pre-training with the framewise CE objective (PT-CE)}\label{ssec:pretrain}
Next, we propose a pre-training of the encoder with framewise CE objective.
Specifically, we first train the encoder only until convergence by setting $\lambda_{\rm CE}$ in Eq. \eqref{eq:mtl} to 1.0 (\textit{stage-1}), and then optimize the entire parameters except for the CE branch by setting $\lambda_{\rm CE}$ to 0 (\textit{stage-2}).
By doing this, we do not have to carefully tune the weight for the framewise CE objective $\lambda_{\rm CE}$.
In this method, we do not stack any linear projection layers on the encoder as in Section \ref{ssec:mtl_frame_xe}.

% system overview

\vspace{-1mm}
\subsection{Leveraging hard alignments in the decoder}
\vspace{-1mm}
%%% delay constrained training
\subsubsection{Delay constrained training (DeCoT)}\label{ssec:decot}
The above two methods utilize the hard alignments on the encoder side.
Here, we leverage them on the decoder side.
Since $\alpha_{i,j}$ is optimized by marginalizing all possible alignments during training, they can include arbitrary future contexts as long as the monotonicity is not violated, which leads to increasing the latency.
Therefore, we remove inappropriate alignments whose boundaries surpass the acceptable latency $\delta$ [frame] in Eq. \eqref{eq:mocha_alpha} as follows:
\begin{eqnarray*}\label{eq:delay_constrained}
\alpha_{i,j} = \begin{cases}
     p_{i,j}\bigg((1-p_{i,j-1})\frac{\alpha_{i,j-1}}{p_{i,j-1}}+\alpha_{i-1,j}\bigg) & (j \le {\rm b}_{i} + \delta) \\
     0 & ({\rm otherwise})
\end{cases}
\end{eqnarray*}
where ${\rm b}_{i}$ is the $i$-th gold boundary.
This delay constrained training (DeCoT) is illustrated in the top left box in Figure \ref{fig:system_overview}.

DeCoT is investigated for the CTC acoustic models in \cite{sak2015acoustic_asru} and we extend it to MoChA, which is also optimized by marginalizing all possible alignment paths.
Unlike CTC, where alignments are calculated with the forward-backward algorithm, the expected boundaries $\{\alpha_{i,j}\}$ in MoChA are calculated only in a forward direction because of the sequential dependency in the decoder.
This causes the exponential decay of $\{\alpha_{i,j}\}$ and leads to almost zero context vectors, especially in the latter part of the output sequence.
To recover this, we introduce a regularization term to keep the number of boundaries as close as possible to the length of output tokens $L$:
\vspace{-1mm}
\[\label{eq:quantity_loss}
\mathcal{L}_{\rm total} = \mathcal{L}_{\rm S2S} + \lambda_{\rm QUA} |L - \sum_{i=1}^{L}{\sum_{j=1}^{T}{\alpha_{i,j}}}|
\]
where $\lambda_{\rm QUA}$ ($\ge 0$) is a tunable hyperparameter.
We name this \textit{quantity loss} inspired by \cite{cif}.
Quantity loss emphasizes the valid alignments during the marginalization process and has the similar effect to re-normalizing of attention weights.
Note that $\{\alpha_{i,j}\}$ should not be explicitly normalized over the encoder outputs ${\bm h}$ since we cannot see the entire outputs during the inference \cite{monotonic_attention}.

%%% Minimum latency training
\subsubsection{Minimum latency training (MinLT)}\label{ssec:minlt}
\vspace{-0.5mm}
The above DeCoT assumes the fixed latency for each token by setting the tolerance $\delta$ to a constant value.
However, the actual latency differs token by token depending on various factors such as speaking speed and the length of characters in each subword.
Therefore, we next explore to directly minimize the expected latency over the target sequence during training.
MinLT is investigated in simultaneous NMT to reduce the expected latency without any supervisions by considering the ratio of input and output lengths \cite{milk_nmt}.
The speeds of reading source tokens and writing target tokens are assumed to be almost constant during the whole translation process.
However, this cannot be directly applied to the ASR task since non-silence frames, which are typically skipped by the decoder, are not uniformly distributed over the input speech.
% In other words, it is impossible to assume that the ratio of the encoding and decoding speeds is constant.
Hence, we design a differentiable expected latency objective for the ASR task and directly minimize it jointly with CE objective $\mathcal{L}_{\rm S2S}$:
% $\mathcal{L}_{\rm MinLT}$
\vspace{-1mm}
\begin{equation}\label{eq:minlt}
\mathcal{L}_{\rm total}=\mathcal{L}_{\rm S2S}+\lambda_{\rm MinLT}\frac{1}{L}\sum_{i=1}^{L}{|\sum_{j=1}^{T}{j\alpha_{i,j}} - {\rm b}_{i}|}
\end{equation}
where $\sum_{j=1}^{T}{j\alpha_{i,j}}$ represents the expected boundary location of the $i$-th token and $\lambda_{\rm MinLT}$ ($\ge 0$) is a tunable hyperparameter.
% We do not use L1 loss because trigger points are sometimes activated before the reference boundaries.
% By jointly optimizing both objectives, we can balance the accuracy and latency at the same time.

% \subsubsection{End-to-end (E2E) boundary shift training}
% Furthermore, we modify the objective function of minimum latency training so that the model can be trained without hard alignments.
% The expected boundary location is directly shifted to the left side by minimizing:
% \begin{eqnarray*}
% \mathcal{L}_{\rm total}=\mathcal{L}_{\rm S2S}+\lambda_{\rm e2e}\frac{1}{|{\bm y}|}\sum_{i=1}^{L}{\sum_{j=1}^{T}{j\alpha_{i,j}}}
% \end{eqnarray*}
% where $\lambda_{\rm e2e}$ ($\ge 0$) is a tunable hyperparameter.
% We set both $\lambda_{\rm MinLT}$ and $\lambda_{\rm e2e}$ to 1.0 in this work.

\vspace{-1mm}
%%%%%%%%%%%%%%%%%%%%%%%%%%%%%%%%%%%%%%%%%%%%%%%%%%
% Experiments
%%%%%%%%%%%%%%%%%%%%%%%%%%%%%%%%%%%%%%%%%%%%%%%%%%
\section{Experiments}\label{sec:exp}
\vspace{-2mm}
\subsection{Experimental conditions}\label{ssec:setting}
\vspace{-0.5mm}
% data
All experiments were conducted on Microsoft's Cortana voice assistant task.
The training data contains around 3.3 million utterances (3.4k hours) in US English. 
The test set contains about 5600 utterances (6 hours).
% Disjoint 4k utterances were extracted from the training data to construct the validation set.
All the data is anonymized with personally identifiable information removed.
% feature extraction
We used 80-channel log-mel filterbank coefficients computed with a 25ms window size and shifted every 10 ms.
Three successive frames were stacked together to form the 240-dimension input features, which results in the time reduction by a factor of 3 (30ms per frame).
% architecture
The encoder consists of 6-layer unidirectional gated recurrent unit (GRU) \cite{gru} with 1024 hidden units in each layer.
The decoders of both offline and streaming S2S models were composed of 2-layer GRU with 512 units per layer.
We performed layer normalization \cite{layer_normalization} after each layer both in the encoder and decoder to stabilize training.
We used chunk size $w=4$ for MoChA.
% among $\{1,2,4,8\}$
% regularization
We used dropout regularization and label smoothing \cite{label_smoothing} with probability 0.1 and 0.2, respectively.
We used the 34k mixed units for the output vocabulary \cite{li2018advancing}.
% optimization
Training was performed using Adam optimizer \cite{adam} with learning rate $2.0 \times 10^{-4}$ and $1.3 \times 10^{-4}$ for the global attention and MoChA, respectively.
We set both $\lambda_{\rm QUA}$ and $\lambda_{\rm MinLT}$ to 1.0.
% decoding
Beam search decoding was performed with beam width 8.
We did not use the external language model for decoding.
% evaluation metric
We report word error rate (WER) on the test set and latency statistics on the validation set since we do not have alignments for the test set.
% TODO: describe the output unit: 34k mixed units

\vspace{-2mm}
\subsection{Results}\label{ssec:result}
\vspace{-1mm}
\subsubsection{Baseline streaming S2S model}
\vspace{-0.5mm}
We first compare the offline and streaming S2S models in Table \ref{tab:result_baseline}.
There are large gaps between the bi- and uni-directional encoders, and also the offline and streaming S2S models.
To try and bridge these gaps, the CTC objective \cite{hybrid_ctc_attention}, external language model integration \cite{google_sota_asr,shallow_fusion,toshniwal2018comparison}, pre-training \cite{adaptive_mocha}, and sequence training such as MBR \cite{mbr_training} can be leveraged.
However, we do not want to make that the focus of this work.
In our experiments, we found that MoChA is very sensitive to hyperparameters such as learning rate.
Tuning the clipping value in Eq. \eqref{eq:mocha_alpha} was also critical to avoid numerical instabilities in our experiments \cite{monotonic_attention}.
% attention masking.
We confirmed 4.24\% relative gain on the original MoChA with the proposed 1-dimensional convolutional (1D-Conv) layer.
Therefore, we use the MoChA with 1D-Conv layer as our baseline in the following experiments.

% baseline
\begin{table}[h]
    \vspace{-5mm}
    \centering
    \footnotesize
    \begingroup
    \caption{Results of the offline and streaming S2S models}\label{tab:result_baseline}
    \vspace{0.5mm}
    \begin{tabular}{|l|l|c|} \hline
      \multicolumn{2}{|c|}{Model} & WER $[$\%$]$ \\ \hline
      \multirow{2}{*}{Offline} & Bidirectional (global attention) & 7.01 \\
      & Unidirectional (global attention) & 8.44 \\ \hline
      \multirow{2}{*}{Streaming} & MoChA (chunk: 4) & 10.37 \\  
      & \ + 1D-Conv (baseline) & {\bf 9.93} \\ \hline
    \end{tabular}
    \vspace{-4.5mm}
    \endgroup
\end{table}

\vspace{-1.5mm}
\subsubsection{Leveraging hard alignments in the encoder}\label{ssec:result_encoder}
\vspace{-0.5mm}
Next, we show results of leveraging hard alignments in the encoder of MoChA in Table \ref{tab:result_encoder}.
With multi-task learning with framewise CE objective (MTL-CE), the latency was significantly reduced at the cost of WER.
As increasing the weight for framewise CE objective $\lambda_{\rm CE}$, further latency reduction was obtained while hurting WER more.
With $\lambda_{\rm CE}=0.3$, we got 40\% latency reduction (median) only with 5.6\% relative WER degradation. 
Pre-training with framewise CE objective (PT-CE) also reduced the latency but sacrificed more WER than MTL-CE.
These observations are contrary to the previous works leveraging framewise CE objective in the CTC model \cite{sak2015acoustic_asru,dfsmn_ctc,multistage_a2w}.
One possible explanation is that CTC is a frame-synchronous model while MoChA is a label-synchronous model, and we did not use phoneme-level supervision to avoid that accuracy gains come from joint optimization with lower-level labels \cite{garg2019improved,toshniwal2017multitask,direct_a2w_is17}.
% Ramon paper

% encoder side
\begin{table}[h]
    \centering
    \footnotesize
    \begingroup
    \vspace{-4mm}
    \caption{Results of latency reduction strategies \underline{on the encoder side}. {\bf 1 frame corresponds to 30ms latency.}}
    \vspace{0.5mm}
    \label{tab:result_encoder}
    \begin{tabular}{|l|c|cccc|} \hline
      \multirow{2}{*}{Model} & WER & \multicolumn{4}{|c|}{Corpus-level latency [frame] ($\downarrow$)} \\ \cline{3-6}
      & [\%] & Ave. & Mid. & 90th & 99th \\ \hline
      Baseline
        & {\bf 9.93} & 11.65 & 10.00 & 21.39 & {\bf 44.29} \\ \hline
      MTL-CE ($\lambda_{\rm CE}=0.1$) 
        & 10.21 & 9.84 & 8.00 & {\bf 19.42} & 46.54 \\  
      MTL-CE ($\lambda_{\rm CE}=0.3$) 
        & 10.48 & 8.78 & 6.00 & 19.69 & 47.96 \\  
      MTL-CE ($\lambda_{\rm CE}=0.5$) 
        & 11.11 & {\bf 8.36} & {\bf 5.00} & 21.21 & 49.86 \\  \hline
      PT-CE
        & 12.74 & 10.49	& 7.00 & 22.90 & 48.65  \\ \hline
    \end{tabular}
    \vspace{-4.5mm}
    \endgroup
\end{table}

\vspace{-1.5mm}
\subsubsection{Leveraging hard alignments in the decoder}\label{ssec:result_decoder}
\begin{figure}[h]
    \centering
    \vspace{-4mm}
    \includegraphics[width=0.95\hsize]{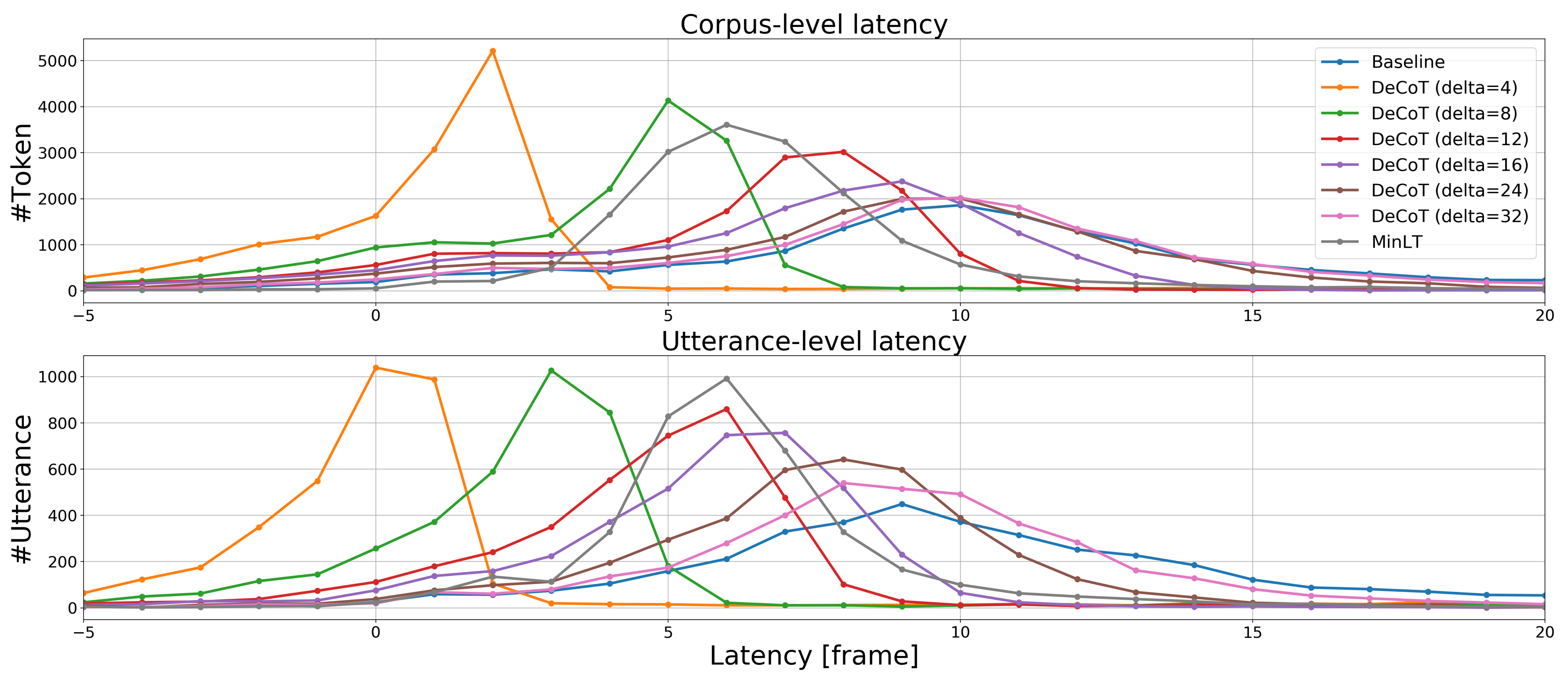}
    \vspace{-4mm}
    \caption{Latency distributions on the decoder side. Top: corpus-level latency $\Delta_{\rm corpus}$, bottom: utterance-level latency $\Delta_{\rm utterance}$.}
    % (top: \textit{corpus-level}, bottom: \textit{utterance-level})
    \label{fig:distribution}
    \vspace{-6mm}
\end{figure}

\vspace{-0.5mm}
We then leverage the hard alignments on the decoder side.
Results are shown in Table \ref{tab:result_decoder} and Figure \ref{fig:distribution}.
We initialized all models expect for the baseline with the baseline MoChA (\textit{warm start}) since alignment constraints in the decoder makes training MoChA from scratch much harder.
Note that we did not provide framewise supervision to the encoder in these experiments.
Both DeCoT and MinLT significantly reduced the latency.
An interesting observation was that WER also improved significantly at the same time with DeCoT with $\delta\ge16$.
We obtained 8.0\% and 10.6\% relative WER improvements with $\delta=16$ and $24$, respectively.
One possible explanation is that clean paths were emphasized more when out-of-boundary paths (potentially noisy) were removed during training.
DeCoT has the effect to reduce the outlier as confirmed from the drastic latency reduction in the tail parts, but too much constraint with the small $\delta$ collapsed the model.
% The reason why latency in the tail parts got much worse is that the exponential decay of $\alpha_{i,j}$ led to skipping boundary predictions and increasing deletion errors.
In contrast, MinLT is effective for moving the center of latency distributions to the left side since the median improved by 40\%.
Note that when the $i$-th boundary corresponding to a non-EOS token is not activated until the last input timestep $T$ (i.e., $\alpha_{i,j}<0.5$), we set the boundary to $T$ to calculate the latency.

Considering the fact that both the latency and WER improved, hard alignments can be used more efficiently on the decoder side.
Since error signals regarding latency were directly connected to the decoder side, the model could balance the accuracy and latency more effectively than techniques on the encoder side.
% the decoder can be aware of error signals of both CE and latency,

% decoder side
\begin{table}[t]
    \centering
    \footnotesize
    \begingroup
    \vspace{-3mm}
    \caption{Results of latency reduction strategies \underline{on the decoder side}}\label{tab:result_decoder}
    \vspace{0.5mm}
    \begin{tabular}{|l|c|cccc|} \hline
      \multirow{2}{*}{Model} & WER & \multicolumn{4}{|c|}{Corpus-level latency [frame] ($\downarrow$)} \\ \cline{3-6}
      & [\%] & Ave. & Mid. & 90th & 99th \\ \hline
    Baseline & 9.93 & 11.65 & 10.00 & 21.39 & 44.29 \\ \hline
      DeCoT ($\delta=4$) & 20.25 & {\bf 3.66} & {\bf 1.00} & 9.56 & 62.27 \\
      DeCoT ($\delta=8$) & 14.35 & 4.60 & 5.00 & {\bf 7.00} & 47.04 \\
      DeCoT ($\delta=12$) & 11.40 & 6.02 & 7.00 & 9.92 & 35.58 \\  
      DeCoT ($\delta=16$) & 9.13 & 6.63 & 8.00 & 11.71 & {\bf 16.43} \\  
      DeCoT ($\delta=24$) & {\bf 8.87} & 8.37	& 9.00 & 14.45 & 21.07 \\
      DeCoT ($\delta=32$) & 9.17 & 9.79 & 10.00 & 16.54 & 27.01 \\ \hline
      MinLT & {\bf 9.70} & {\bf 7.06} & {\bf 6.00} & {\bf 10.63} & {\bf 26.76} \\ \hline
    \end{tabular}
    \vspace{-5mm}
    \endgroup
\end{table}

Finally, we conducted the ablation study on the decoder side in Table \ref{tab:result_ablation}.
Quantity loss was not necessary for the baseline MoChA and MinLT, but essential for DeCoT.
Warm start training was also necessary for both DeCoT and MinLT.
This is probably because incorrect error signals were propagated into the model in the early training stage when training from scratch.
The combination of DeCoT and MinLT with warm start training degraded WER too much although the latency was reduced further.
The baseline WER was indeed boosted thanks to more updates by warm start training at the cost of latency, but DeCoT with $\delta=16, 24$ were still better than it with much smaller latency.
We also tried to directly shift the boundary locations without boundary supervisions by setting $b_{i}$ to 0 for all $i$ in Eq. \eqref{eq:minlt}.
However, this did not lead to the latency reduction, from which we can confirm the effectiveness of our proposed expected latency objective.

% ablation study
\begin{table}[h]
    \centering
    \footnotesize
    \begingroup
    \vspace{-4mm}
    \caption{Ablation study on the decoder side}\label{tab:result_ablation}
    \vspace{0.5mm}
    \begin{tabular}{|l|c|cccc|} \hline
      \multirow{2}{*}{Model} & WER & \multicolumn{4}{|c|}{Corpus-level latency [frame] ($\downarrow$)} \\ \cline{3-6}
      & [\%] & Ave. & Mid. & 90th & 99th \\ \hline
      Baseline & 9.93 & 11.65 & 10.00 & 21.39 & 44.29 \\
      \ \ w/ warm start & {\bf 9.21} & 12.27 & 11.00 & 22.23 & 43.16 \\
      \ \ w/ quantity loss & 10.30 & 11.24 & 10.00 & 20.39 & 36.01 \\ \hline
      DeCoT ($\delta=16$) & {\bf 9.13} & 6.63 & 8.00 & 11.71 & 16.43 \\  
      \ \ w/o warm start & 10.72 & 6.28 & 7.00 & 11.12 & 36.03  \\ 
      \ \ w/o quantity loss & 14.28 & 3.93 & 3.00 & 7.20 & 27.39  \\  
      \ \ w/ \ \ MinLT & 12.75 & {\bf 4.05} & {\bf 4.00} & {\bf 7.96} & {\bf 15.92} \\ \hline
      MinLT & {\bf 9.70} & 7.06 & 6.00 & 10.63 & 26.76 \\ 
      \ \ w/o warm start & 13.60 & 11.83 & 10.00 & 21.41 & 45.06 \\ 
      \ \ w/ \ \ quantity loss & 13.66 & 6.82 & 6.00 & 10.45 & 25.57  \\ 
      \ \ w/ \ \ $b_{i}=0$ in Eq. \eqref{eq:minlt} & 9.29 & 12.11 & 11.00 & 21.77 & 42.85 \\ 
      \hline
    \end{tabular}
    \vspace{-2mm}
    \endgroup
\end{table}

% curriculum
% \begin{table}[h]
%     \centering
%     \begingroup
%     \footnotesize
%     \begin{tabular}{|l|c|cccc|} \hline
%       \multirow{2}{*}{Model} & \multirow{2}{*}{\shortstack{WER}} & \multicolumn{4}{|c|}{Corpus-level latency [frame]} \\ \cline{3-6}
%       &  & Ave. & Mid. & 90\%ile & 99\%ile \\ \hline
%       MinLT from DeCoT ($\delta=16$) & run &   &   &   &   \\
%       MinLT from DeCoT ($\delta=24$) & run &   &   &   &   \\ \hline
%     %   DeCoT ($\delta=16$) from MinLT & run &   &   &   &   \\
%     %   DeCoT ($\delta=24$) from MinLT & run &   &   &   &   \\ \hline
%     \end{tabular}
%     \vspace{-3mm}
%     \caption{Curriculum learning}\label{tab:result_curriculum}
%     \vspace{-3mm}
%     \endgroup
% \end{table}

\vspace{-0.5mm}
%%%%%%%%%%%%%%%%%%%%%%%%%%%%%%%%%%%%%%%%%%%%%%%%%%
% Conclusion
%%%%%%%%%%%%%%%%%%%%%%%%%%%%%%%%%%%%%%%%%%%%%%%%%%
\section{Conclusion}\label{sec:conclusion}
\vspace{-2mm}
In this paper, we tackled the delayed token generation problem for the streaming attention-based S2S ASR model.
We explored to leverage external hard alignments obtained from the hybrid ASR model to make the decision for the next token generation as fast as possible while maintaining the recognition accuracy.
We proposed several strategies which are applicable to the encoder and decoder subnetworks.
% to reduce the latency..
Experimental evaluation demonstrated that hard alignments were effective in both subnetworks for latency reduction and further reduced word error rate when applied to the decoder.
% For future work, we will replace these hard alignments with those obtained from the offline S2S model.
% RNN-T

% \newpage
\footnotesize
\bibliographystyle{IEEEbib}
\bibliography{reference}

\end{document}